\documentclass[letterpaper]{article}
\usepackage{aaai16}
\usepackage{times}
\usepackage{helvet}
\usepackage{courier}
\usepackage{amsmath,amsthm}

\newtheorem{de}{Definition}
\newtheorem{thm}{Theorem}

\newtheorem{pro}{Proposition}

\newcommand{\comment}[1]{}
\newcommand{\mc}{\mathcal}

\def\dm{\leavevmode\baselineskip0pt \setbox1=\hbox{$-$}
   \setbox2=\hbox to \wd1{\hfil.\hfil}
   \vbox{\box2\vskip-0.3ex\box1}}

\title{A New Approach for Revising Logic Programs}
\author{Zhiqiang Zhuang$^{1}$\; James Delgrande$^{2}$ \; Abhaya Nayak$^{3}$ \; Abdul Sattar$^{1}$\\
$^{1}$ Institute for Integrated and Intelligent Systems, 
       Griffith University, Australia \\
$^{2}$ School of Computing Science, Simon Fraser University, Canada \\
$^{3}$ Department of Computing, Macquarie University, Australia}

\begin{document}
\maketitle

\begin{abstract}

Belief revision has been studied mainly with respect to background logics that are monotonic in character.
In this paper we study belief revision when the underlying logic is non-monotonic
instead---an inherently interesting problem that is under explored.
In particular, we will focus on the revision of a body of beliefs 
that is represented as a logic program under the answer set semantics, while
the new information is also similarly represented as a logic program.
Our approach is driven by the observation that unlike in a monotonic setting where, 
when necessary,  consistency in a revised body of beliefs is maintained
 by jettisoning some old beliefs,
in a non-monotonic setting consistency can be restored by adding new beliefs as well.
We will define a syntactic revision function
and subsequently provide representation theorem for characterising it.


\end{abstract}

\section{Introduction}



The ability to change one's beliefs when presented with new information
is crucial for any intelligent agent.
In the area of \textit{belief change}, substantial effort has been made
towards the understanding and realisation of this process.
Traditionally, it is assumed that the agent's reasoning is governed by a
monotonic logic.
For this reason, traditional belief change is inapplicable 
when the agent's reasoning is non-monotonic.
Our goal in this research program is to extend
belief base \cite{book/Hansson99} approaches in belief revision 
to nonmonotonic setting. 
In this paper, we focus on \textit{disjunctive logic programs}, as a well-studied and well-known approach to nonmonotonic reasoning that also has efficient implementations.

Much, if not most, of our day-to-day reasoning involves non-monotonic reasoning.
To illustrate issues that may arise, consider the following example.
In a university, professors generally teach, unless they have an
administrative appointment.
Assume we know that John is a professor. 
Since most faculty do not have an administrative appointment, and there is no
evidence that John does, we conclude that he teaches. 
This reasoning is a classical form of non-monotonic reasoning, namely using the
\textit{closed world assumption}.
It can be represented by the
following logic program under the \emph{answer set semantics}.
\begin{align}
Teach(X) &\leftarrow Prof(X), not\; Admin(X).\\
Prof(John) &\leftarrow. 
\end{align}
The \textit{answer set} $\{Prof(John), Teach(John)\}$ for this logic program  
corresponds exactly to the facts we can conclude.

Suppose we receive information that John does not teach, which we can
represent by the rule
\begin{align}
\leftarrow Teach(John).
\end{align}
Now our beliefs about John are contradictory; and it is not surprising that 
the logic program consisting of rules (1) -- (3) has no answer set.
For us or any intelligent agent in this situation to function properly,
we need a mechanism to resolve this inconsistency.
This is a typical belief revision problem;
however, the classical (AGM) approach can not be applied, 
as we are reasoning non-monotonically.

It is not hard to suggest possible causes 
of the inconsistency and to resolve it.
It could be that some of our beliefs are wrong;
perhaps professors with administrative duties may still need to do teaching or
perhaps John is not a professor. 
Thus we can restore consistency by removing rule (1) or (2).
Alternatively and perhaps more interestingly, 
it could be that assuming that John is not an administrative staff
via the absence of evidence is too adventurous;
that is he may indeed be an administrative staff member but we don't know it.
Thus we can also restore consistency by adding the missing evidence of John
being an administrative staff member by
\begin{align}
Admin(John) &\leftarrow.
\end{align}

The second alternative highlights the distinction for belief revision
in monotonic and non-monotonic settings. 
In the monotonic setting, an inconsistent body of knowledge will 
remain inconsistent no matter how much extra information is supplied.
On the other hand, in the non-monotonic setting, inconsistency
can be resolved by either removing old information, 
or adding new information, or both.
Therefore, belief revision functions 
in a non-monotonic setting 
should allow a mixture of removal and addition of information
for inconsistency-resolution.
In this paper, we will define one such 
revision functions for disjunctive logic programs 
under the answer set semantics.

The revision function is called \textit{slp-revision} and
is a belief base revision which
takes syntactic information into account.
In revising $P$ by $Q$,
an slp-revision function first obtains
a logic program $R$ that is consistent with $Q$
and differs minimally from $P$, then combines $R$ with $Q$.
For example, if $P=\{ (1), (2)\}$ and $Q=\{(3)\}$, 
then $R$ could be $\{(1)\}$ (i.e., resolving inconsistency by removing (2)); 
$\{(2)\}$ (i.e., resolving inconsistency by removing (1)); or $\{(1), (2), (4)\}$ (i.e., resolving inconsistency by adding $(4)$).

 The next section gives logical preliminaries. The following one develop 
 our approach to slp-revision in which we provide postulates, 
 a semantic construction, and a representation result. 
This is followed by a comparison to other work, and a brief conclusion.

\section{Preliminary Considerations}

In this paper, we consider only fully grounded disjunctive logic programs.
That is variables in program rules 
are replaced by the set of their ground instances.
Thus a logic program (or program for short) here 
is a finite set of rules of the form:
$$a_1;\ldots; a_m\leftarrow b_1,\ldots, b_n, not\; c_{1},\ldots, not\; c_{o}$$ 
where $m,n,o \geq 0$, $m+n+o>0$, and
$a_i,b_j,c_k\in \mathcal{A}$ for $\mathcal{A}$ a finite set of propositional atoms.
Connective \textit{not} is called \textit{default negation}.
We denote the set of all logic programs by $\mc{P}$.
For each rule $r$, 
let $H(r)=\{a_1,\ldots,a_n\}$,
$B^+(r)=\{b_1,\ldots,b_m\}$,
and $B^-(r)=\{c_1,\ldots, c_o\}$.
The letters $P$ and $Q$ are used to denote a logic program
throughout the paper. 

An interpretation is represented by the subset of atoms in $\mc{A}$ 
that are true in the interpretation.
A \textit{classical model} of a program $P$ is an 
interpretation in which all rules of $P$ are true according to 
the standard definition of truth in propositional logic, and
where default negation is treated as classical negation.
The set of classical models of $P$ is denoted as $Mod(P)$. 
Given an interpretation $Y$, we write $Y\models P$ to mean
$Y$ is a classical model of $P$.
The \textit{reduct} of a program $P$ with respect to an interpretation $Y$,
denoted $P^Y$, is the set of rules:
$$\{H(r)\leftarrow B^+(r)\,|\, r\in P, B^-(r)\cap Y=\emptyset\}.$$
An \textit{answer set} $Y$ of $P$ is a subset-minimal classical model of $P^Y$.
The set of all answer set of $P$ is denoted as $AS(P)$.

An \textit{SE interpretation} \cite{journals/tplp/Turner03} 
is a pair $(X,Y)$ of interpretations
such that $X\subseteq Y\subseteq \mc{A}$.
The set of all SE interpretations (over $\mc{A}$) is denoted $\mc{SE}$.
The letters $M$ and $N$ are used to denote a set of SE interpretations
throughout the paper.
An SE interpretation is an \textit{SE model} of a program $P$
if $Y\models P$ and $X\models P^Y$.
The set of all SE models of $P$ is denoted as $SE(P)$.
SE models are proposed to capture
\textit{strong equivalence} \cite{journals/tcl/Lifschitz01} between programs
that is $SE(P)=SE(Q)$ iff $P$ and $Q$ are strongly equivalent,
thus they contain more informations than answer sets.

The following two properties of SE models \cite{journals/tplp/Turner03} 
are crucial to this paper:
\begin{enumerate}
\item  $Y\in AS(P)$ iff $(Y,Y)\in SE(P)$ and there is no $(X,Y)\in SE(P)$ 
such that $X\subset Y$.
\item $(Y,Y)\in SE(P)$ iff $Y\in Mod(P)$.
\end{enumerate}
So $SE(P)\neq\emptyset$ iff  $Mod(P)\neq\emptyset$
but $SE(P)\neq\emptyset$ does not imply $AS(P)\neq\emptyset$.
This gives rise to two notions of consistency.

\begin{de}
$P$ is \emph{consistent} iff $AS(P)\neq \emptyset$
and $P$ is \emph{m-consistent}\footnote{``m'' stands for ``monotonic'' which indicates 
that the notion of m-consistency is based on a monotonic characterisation (i.e., SE models) for logic programs.} iff $SE(P)\neq\emptyset$.
\end{de}
\noindent
It is clear from the SE model properties that 
consistency implies m-consistency; m-inconsistency implies inconsistency.
In other words, a consistent program is m-consistent but not vice versa.

In subsequent sections, 
we will need to describe the difference between two logic programs.
For this purpose, 
we use the \textit{symmetric difference} operator $\ominus$ which is 
defined as
$$X \ominus Y=(X\setminus Y) \cup (Y\setminus X)$$
for any sets $X$ and $Y$.

\section{SLP-Revision Functions}



In this section, we give a syntax-based revision function
$*:\mc{P}\times\mc{P}\mapsto\mc{P}$
 for revising one logic program by another.
The function takes a logic program $P$ called the \textit{original logic program} 
and a logic program $Q$ called the \textit{revising logic program}, 
and returns another logic program $P*Q$ called the \textit{revised logic program}.
Following AGM belief revision, we want to have
$Q$ contained in $P*Q$ (i.e., $Q\subseteq P*Q$),
$P*Q$ is consistent whenever possible, and 
that as much of $P$ as consistently possible is contained in $P*Q$.

Clearly, a key issue in defining $*$ is to deal with 
the possible inconsistency between $Q$ and $P$.
As illustrated in the teaching example, one means of ensuring that $P*Q$ is
consistent is to remove a minimal set of beliefs from $P$ so that adding $Q$
to the result is consistent. Of course there may be more than one way to 
remove beliefs from $P$.
Following this intuition,
we obtain all maximal subsets of $P$ that are consistent with $Q$, which we
call the \emph{s-removal compatible programs} 
of $P$ with respect to $Q$.

\begin{de}
The set of \emph{s-removal compatible programs} of $P$ with respect to $Q$, denoted $P\downarrow Q$,
is such that $R\in P\downarrow Q$ iff\\
1.\ $R\subseteq P$,\\
2.\ $R \cup Q$ is consistent, and \\
3.\ if $R\subset R'\subseteq P$, then $R' \cup Q$ is inconsistent.
\end{de}

\noindent

The notion of s-removal compatible programs is not new,
classical revision functions 
\cite{journals/jsyml/AlchourronGM85,journals/jpl/Hansson93} 
are based on more or less the same notion.
The difference is that this notion alone is sufficient to capture the 
inconsistency-resolution strategy of classical belief revision, 
but there is more that one can do in non-monotonic belief revision.

In our non-monotonic setting, we are able to express assumptions 
(i.e., negation as failure) and to reason with them.
Earlier, we assumed John is not an administrator, in the
absence of evidence to the contrary.
With this, we came to the conclusion that he has to teach.
Consequently, if we learn that John does not teach, as in our example, one way of resolving this inconsistency is by adding information 
so that our assumption does not hold.
Following this intuition,
we obtain all the minimal supersets of $P$ that are consistent with $Q$, 
which we call the \textit{s-expansion compatible program} 
of $P$ with respect to $Q$.

\begin{de}
The set of \emph{s-expansion compatible programs} of $P$ with respect to $Q$, 
denoted $P\uparrow Q$,
is such that $R\in P\uparrow Q$ iff\\
1.\ $P\subseteq R$,\\
2.\ $R\cup Q$ is consistent, and \\
3.\ if $P\subseteq R'\subset R$, then $R' \cup Q$ is inconsistent.
\end{de}

\noindent

Since the s-expansion and s-removal compatible programs
are consistent with $Q$ and are obtained by removing or adding
minimal sets of rules from or to $P$, 
the union of $Q$ with any of these sets is consistent and comprises
a least change made to $P$ in order to achieve consistency.
These programs clearly should be candidates 
for forming the revised logic program $P*Q$;
however, they do not form the set of all candidates.
In particular, we can obtain a program that differs the least from $P$ and is
consistent with $Q$ by removing some beliefs of $P$ and at the same time 
adding some new beliefs to $P$.
Thus we consider all those logic programs
that differ the least from $P$ and are consistent with $Q$;
these are called the \textit{s-compatible programs} 
of $P$ with respect to $Q$.

\begin{de}
The set of \emph{s-compatible programs} of $P$ with respect to $Q$, denoted $P\updownarrow Q$,
is such that $R\in P\updownarrow Q$ iff\\
1.\ $R\cup Q$ is consistent and\\
2.\ if $P\ominus R' \subset P\ominus R$, then $R'\cup Q$ is inconsistent.
\end{de}

\noindent
For example, let
$P=\{a\leftarrow b, not\,c.,\, b., \,e\leftarrow f, not\,g., \,f.\}$ and 
$Q=\{\leftarrow a., \leftarrow e.\}$.
Then $P\cup Q$ is inconsistent since
$a$ and $e$ can be concluded from $P$ but they contradict the rules of $Q$. 
To resolve the inconsistency via making the least change to $P$,
we could remove $b\leftarrow$ from $P$ 
(which eliminates the contradiction about $a$)
and add $g\leftarrow$ to $P$
(which eliminates the contradiction about $e$).
The program thus obtained (i.e., $(P\setminus \{b.\})\cup \{g.\}$) is  
a s-compatible program in $P\updownarrow Q$.

It is obvious, but worth noting that 
the notion of s-compatible program subsumes those of
s-removal and s-expansion compatible programs.
In the above example, $P\updownarrow Q$ also contains
$P\setminus \{b., f.\}$ and $P\cup \{c., g.\}$, which 
are respectively an s-removal
and an s-expansion compatible program of $P$ with respect to $Q$.

\begin{pro}
$(P\uparrow Q) \cup (P\downarrow Q) \subseteq P\updownarrow Q$.
\end{pro}

There are cases in which we cannot resolve inconsistency
by only adding new beliefs which means
the set of s-expansion compatible programs is empty.
For example, if $P=\{a.\}$ and $Q=\{\leftarrow a.\}$, then
$P\cup Q$ is inconsistent and we cannot restore consistency
without removing $a\leftarrow$ from $P$.
In these cases, the inconsistency is due to
contradictory facts that can be concluded 
without using any reasoning power beyond 
that of classical logic.
Clearly, the inconsistency is of a monotonic nature,
that is, in our terminology, m-inconsistency.


\begin{pro}\label{lem:empty-expansion-set}
If $P\cup Q$ is m-inconsistent, then 
$P\uparrow Q=\emptyset$.
\end{pro}

So far, we have identified the candidates for forming $P*Q$.
It remains to pick the ``best'' one.
Such extralogical information is typically modelled by a
\emph{selection function}, which we do next.

\begin{de}
A function $\gamma$ is a \emph{selection function}
for $P$ iff for any program $Q$,
$\gamma(P\updownarrow Q)$ returns a single element of $P\updownarrow Q$
whenever $P\updownarrow Q$ is non-empty;
otherwise it returns $P$.
\end{de}

\noindent
The revised logic program $P*Q$ is then formed by combining $Q$ with 
the s-compatible program picked by the selection function for $P$.
We call the function $*$ defined in this way 
 a \emph{slp-revision function} for $P$.

\begin{de}
A function $*$ is a slp-revision function for $P$ iff
$$P*Q=\gamma (P\updownarrow Q)\cup Q$$
for any program $Q$, where $\gamma$ is a selection 
function for $P$.
\end{de}

\noindent

In classical belief revision, multiple candidates maybe chosen by a
selection function, and their intersection is combined with the new belief 
to form the revision result.
There, a selection function that picks out a single element is called a \emph{maxichoice} function  \cite{journals/jsyml/AlchourronGM85}.
In classical logic, maxichoice selection functions leads to 
undesirable properties for belief set revision but not for belief base revision.
In our non-monotonic setting, picking
multiple candidates does not make sense, as
intersection of s-compatible programs may
not be consistent with the revising program.
For example, let $P=\{a\leftarrow not\,b, not\,c.\}$
and $Q=\{\leftarrow a.\}$.
We can restore consistency of $P$ with $Q$ 
by, for instance, adding the rule $b\leftarrow$
to $P$ which corresponds to the s-compatible program $P\cup \{b.\}$
or by adding the rule $c\leftarrow$
which corresponds to the s-compatible program $P\cup\{c.\}$.
However, the intersection of the two s-compatible programs is inconsistent with $Q$.

We turn next to properties of slp-revision functions.
Consider the following set of postulates where 
$*:\mc{P}\times\mc{P}\mapsto\mc{P}$ is a function.

\noindent
\begin{tabular}{l l}
(s$*$s)  & \hspace{-4mm} $Q\subseteq P*Q$\\
(s$*$c)  & \hspace{-4mm} If $Q$ is m-consistent, then $P*Q$ is consistent\\
(s$*$f)  & \hspace{-4mm} If $Q$ is m-inconsistent, then $P*Q=P\cup Q$\\
(s$*$rr) & \hspace{-4mm} If $R\neq \emptyset$ and $R\subseteq P\setminus (P*Q)$, then\\
           & \hspace{-4mm} $(P*Q)\cup R$ is inconsistent\\
(s$*$er) & \hspace{-4mm} If $E\neq \emptyset$ and $E\subseteq (P*Q)\setminus (P\cup Q)$, then\\
        & \hspace{-4mm} $(P*Q)\setminus E$ is inconsistent\\
(s$*$mr) &  \hspace{-4mm} If $R\neq \emptyset$, $R\subseteq P\setminus (P*Q)$, \\ 
  & \hspace{-4mm}  $E\neq \emptyset$ and $E\subseteq (P*Q)\setminus (P\cup Q)$, then \\
             & \hspace{-4mm} $((P*Q)\cup R)\setminus E$ is inconsistent\\
(s$*$u)  & \hspace{-4mm} If $P\updownarrow Q=P\updownarrow R$, then\\
          & \hspace{-4mm} $P \setminus (P*Q) = P \setminus (P*R)$ and\\
          & \hspace{-4mm} $(P*Q) \setminus (P\cup Q) = (P*R) \setminus (P\cup R)$  
\end{tabular}


\noindent
(s$*$s) (\textit{Success}) states that a revision is always successful in incorporating 
the new beliefs.
(s$*$c) (\textit{Consistency}) states that 
a revision ensures consistency of the revised logic program whenever possible.
In the monotonic setting, a revision results in inconsistency only when the
new beliefs are themselves inconsistent.
This is not the case in the non-monotonic setting.
For example, consider the revision of $P=\{a.\}$ by $Q=\{b\leftarrow not\;b\}$.
Although $Q$ is inconsistent, we have $P\cup \{b.\}$ as a s-compatible program 
of $P$ with respect to $Q$.
Thus we can have $P \cup\{b.\}\cup Q$ as 
the revised logic program, which contains $Q$ and is consistent.
Here, a revision results in inconsistency 
only when the revising logic program is m-inconsistent.
In such a case, (s$*$f) (\textit{Failure}) states that 
the revision corresponds to the union
of the original and revising logic program.

(s$*$rr) (\textit{Removal Relevance}) states that
if some rules are removed from the original logic program for the revision,
then adding them to the revised logic program results in inconsistency. 
It captures the intuition that nothing is removed unless
its removal contributes to making the revised logic program consistent.
(s$*$er) (\textit{Expansion Relevance}) states that
if some new rules other than those in the revising logic program
are added to the original logic program for the revision,
then removing them from the revised logic program
causes inconsistency. 
It captures the intuition that nothing is added unless
adding it contributes to making the revised logic program consistent.
(s$*$mr) (\textit{Mixed Relevance}) states that if some rules are removed from
the original logic program and some new rules other than those in the revising logic program are added to the original logic program for the revision, 
then adding back the removed ones and removing the added ones result in
inconsistency of the revised logic program.
Its intuition is a mixture of the two above.
Note that putting (s$*$rr) and (s$*$er)
together does not guarantee (s$*$mr), nor the reverse.
In summary, these three postulates express the necessity of
adding and/or removing certain belief for resolving inconsistency 
and hence to accomplish a revision.
In classical belief revision, inconsistency can only be resolved by removing
old beliefs;
the necessity of removing particular beliefs is captured by the 
\textit{Relevance} postulate \cite{journals/jpl/Hansson93}.\footnote{If $\psi\in K$ and
$\psi\not\in K*\phi$, then there is some $K'$ such that $K*\phi\subseteq K'\subseteq K\cup\{\phi\}$, $K'$ is consistent but $K'\cup \{\psi\}$ is inconsistent.}
The three postulates are the counterparts of \textit{Relevance}
in our non-monotonic setting,
and we need all three of them to deal respectively with addition, removal, 
and a mixture of addition and removal.


Finally, (s$*$u) (\textit{Uniformity}) states the condition under which
two revising logic programs $Q$ and $R$
trigger the same changes to the original logic program $P$.
That is the rules removed from $P$ (i.e., $P \setminus (P*Q)$) and
the rules added to $P$ (i.e., $(P*Q) \setminus (P\cup Q)$)
for accommodating $Q$
are identical to those for accommodating $R$.
Certainly having $Q$ and $R$ be strongly equivalent (i.e., $SE(Q)=SE(R)$)
is a sufficient condition.
However, it is too strong a requirement.
Suppose $P=\{\leftarrow a.\}$,
$Q=\{a.\}$, and $R=\{a\leftarrow b., b.\}$. 
Then the minimal change to $P$ we have to made to accommodate 
$Q$ and $R$ are the same, that is we remove $\leftarrow a$.
However $Q$ and $R$ are not strongly equivalent, even though they incur the
same change to $P$.
The essential point of this example is that
instead of a global condition like strong equivalence, 
we need a condition that is local to the original logic program $P$.
Unfortunately, it seems there is no existing notion in the logic programming
literature that captures this local condition.
Thus we use our newly defined notion of s-compatible programs and
come up with the local but more appropriate condition in (s$*$u).

We can show that these postulates are sufficient to characterise
all slp-revision functions.

\begin{thm}\label{thm:slp-revision}
A function $*$ is a slp-revision function iff it satisfies
(s$*$s), (s$*$c), (s$*$f), (s$*$rr), (s$*$er), (s$*$mr), and (s$*$u).
\end{thm}

\section{Comparisons with Existing Approaches}


There has been much work on belief revision for logic programs.
The seminal work of Delgrande et al \shortcite{journals/tocl/DelgrandeSTW13} 
generalises Satoh's \shortcite{conf/Satoh88} and Dalal's \shortcite{conf/aaai/Dalal88}
 revision operators to logic programs. 
Significantly, they bring SE model into the picture.
They do not work with answer sets as a basis for revision, 
 but rather they base their definitions directly on SE models.
The work has inspired several other SE model approaches.
Schwind and Inoue \shortcite{conf/lpnmr/SchwindI13}
provide a constructive characterisation for the revision operators 
in \cite{journals/tocl/DelgrandeSTW13}.
Delgrande et al \shortcite{conf/lpnmr/DelgrandePW13} 
adapt the model-based revision of Katsuno and Mendelzon \shortcite{journals/ai/KatsunoM92}
to logic programs and provide a representation theorem. 
Finally, Binnewies et al \shortcite{conf/aaai/Binnewies15} 
provide a variant of partial meet revision and contraction for logic programs.

Firstly, the SE model approaches are essentially belief set revision
whereas our slp-revision is a belief base one.
Secondly and more importantly,
these approaches assume a weaker notion of consistency, that is m-consistency.
For this reason, some contradictions will not be dealt with in these approaches.
For instance, the contradictory rule $a\leftarrow not\, a$ is m-consistent
thus is considered to be an acceptable state of belief.
Also in our teaching example,
as the program consisting of rules (1) -- (3) is m-consistent,
no attempt will be made to resolve 
the contradiction about John's teaching duty by the SE model approaches.
Therefore for application scenarios in which such contradictions can not be tolerant,
our llp-revision function is clearly a better choice.

Apart from the SE model approaches,
Kr\"umpelmann and Kern-Isberner \shortcite{conf/jelia/KrumpelmannK12} 
provide a revision function for logic programs that originates
from Hansson's \textit{semi-revision} \cite{journals/jancl/Hansson97}.
Since they assume the same notion of consistency as ours, 
all the above mentioned contradictions will be resolved in their approach.

As we have noted, classical belief revision 
is defined for monotonic setting, not for non-monotonic ones.
Inconsistency can be caused by wrong assumptions 
in the non-monotonic setting but not in the monotonic setting.
Such causes are not considered in \cite{conf/jelia/KrumpelmannK12}.
Consequently, their approach only support one of the many possible
inconsistency-resolution strategies we have developed.
Specifically, in \cite{conf/jelia/KrumpelmannK12},
inconsistency can be resolved only by removing old beliefs;
this strategy is captured by a notion 
analogous to s-removal compatible programs.
The inconsistency-resolution strategies captured
by the notion of s-expansion compatible program and 
s-compatible program in general are not considered.

\section{Conclusion and Future Work}

Depending on the application scenario, 
the logic governing an agent's
beliefs could be either monotonic or non-monotonic.
Traditional belief revision assumes that an agent reasons monotonically;
therefore, by definition, it is applicable to such situations only.
Here we have aimed to provide a belief revision framework for situations
in which the agent reasons non-monotonically.
To this end, we defined a belief revision function
for disjunctive logic programs under the answer set semantics. 

Inconsistency-resolution is an essential task for belief revision.
However, the strategies used in traditional belief revision functions
are limited to situations when the agent reasons monotonically.
With a logic program we have the luxury of making assumptions via lack of
contrary evidence, and we can deduce certain facts from such assumptions.
Thus if a set of beliefs is inconsistent, then one possible cause is that we
made the wrong assumption.
In such cases, we can resolve the inconsistency by adding some new rules
so that the assumption can no longer be made.
Such a cause of inconsistency and the associated inconsistency-resolution
strategy is beyond the scope of traditional belief revision, but is crucial 
for non-monotonic belief revision.
We argue that this rationale, which is encoded in our belief revision
function, captures the fundamental difference between monotonic and
non-monotonic belief revision.

This paper then has explored belief base revision 
in the non-monotonic setting of disjunctive logic programs.
Note that the characterising postulates of the base revision 
are formulated in terms of set-theoretic notions (e.g., subsets, set
differences);
the only logical notion required is consistency.
Moreover the key idea, namely the notion of s-compatible programs,
is also based on the same set-theoretic and logical notions.
These notions are present in all non-monotonic settings.
In future work we propose to extend the base revision 
to a general approach to belief revision 
in arbitrary non-monotonic settings.


\section{Appendix: Proof of Results}

In this appendix, we give the proof for the main results.\\

\noindent
\textbf{Proof for Proposition~\ref{lem:empty-expansion-set}}

Let $P$ and $Q$ are logic programs.
Suppose $P\cup Q$ is m-inconsistent. 
We need to show $P\uparrow Q=\emptyset$.

Since $P\cup Q$ is m-inconsistent, 
we have $SE(P)\cap SE(Q)=\emptyset$.
By the definition of s-expansion compatible program, 
any element in $P\uparrow Q$ has to be a superset of $P$ and consistent with $Q$.
However, for any superset $R$ of $P$, 
we have $SE(R)\subseteq SE(P)$.
Thus $SE(R)\cap SE(Q)=\emptyset$ which implies $R\cup Q$ is m-inconsistent.

\qed

\noindent
\textbf{Proof for Theorem~\ref{thm:slp-revision}}

For one direction, suppose $*$ is a slp-revision function for $P$ 
and the associated selection function is $\gamma$. 
We need to show $*$ satisfies 
(s$*$s), (s$*$c), (s$*$f), (s$*$rr), (s$*$er), (s$*$mr), and (s$*$u).
(s$*$s), (s$*$c), and (s$*$f)
follow immediately from the definition of slp-revision functions and compatible programs.

(s$*$rr): 
Suppose there is a set $R$ such that $R\neq \emptyset$ and 
$R\subseteq P\setminus (P*Q)$.
By the definition of slp-revision, 
we have $P*Q=\gamma(P\updownarrow Q)\cup Q$, hence 
$P\setminus (\gamma(P\updownarrow Q)\cup Q)\neq \emptyset$ which implies 
$\gamma(P\updownarrow Q)\neq P$. Then it follows from 
the definition of selection function that $P\updownarrow Q\neq \emptyset$
and $\gamma(P\updownarrow Q)\in P\updownarrow Q$.
Let $\gamma(P\updownarrow Q)=X$.
Then $(P*Q)\cup R=X\cup Q\cup R$.
Since $\emptyset\not=R\subseteq P$,  we have
$((X\cup R) \ominus P) \subset (X\ominus P)$.
By the definition of compatible program, $X\cup R\cup Q$ is inconsistent
that is $(P*Q)\cup R$ is inconsistent.

(s$*$er): 
Suppose there is a set $E$ such that $E\neq \emptyset$ and 
$E\subseteq (P*Q)\setminus (P\cup Q)$.
By the definition of slp-revision, 
we have $P*Q=\gamma(P\updownarrow Q)\cup Q$, hence 
$(\gamma(P\updownarrow Q)\cup Q)\setminus (P\cup Q) \neq \emptyset$ which implies 
$\gamma(P\updownarrow Q)\neq P$. Then it follows from 
the definition of selection function that $P\updownarrow Q\neq \emptyset$
and $\gamma(P\updownarrow Q)\in P\updownarrow Q$.
Let $\gamma(P\updownarrow Q)=X$.
Then $(P*Q)\setminus E=(X\cup Q)\setminus E$.
Since $E\cap P=\emptyset$ and 
$\emptyset\neq E\subseteq X$,
$((X\setminus E) \ominus P) \subset (X\ominus P)$.
By the definition of compatible program, $(X\setminus E)\cup Q$ is inconsistent.
Then since $E\cap Q=\emptyset$, we have
$(X\setminus E)\cup Q= (X\cup Q)\setminus E=(P*Q)\setminus E$.
Thus $(P*Q)\setminus E$ is inconsistent.
 
(s$*$mr): Can be proved by combining the proving method for (s$*$rr) and (s$*$er).

(s$*$u):
Suppose $P\updownarrow Q=P\updownarrow R$. 
Then $\gamma(P\updownarrow Q)=\gamma(P\updownarrow R)$.
If $P\updownarrow Q=P\updownarrow R=\emptyset$, 
then by the definition of slp-revision
$P*Q=P\cup Q$ and $P*R=P\cup R$.
Thus $P\setminus (P*Q)=P\setminus (P*R)=\emptyset$
and $(P*Q)\setminus (P\cup Q)=(P*R)\setminus (P\cup R)=\emptyset$.
So suppose $P\updownarrow Q=P\updownarrow R\neq \emptyset$
and let $X=\gamma(P\updownarrow Q)=\gamma(P\updownarrow R)$.
By the definition of slp-revision, we have 
$P\setminus (P*Q)=P\setminus (X\cup Q)$.
Assume $\emptyset\neq P\cap Q\not\subseteq X$.
Then since $X \cup (P\cap Q)$ is consistent with $Q$
and $(X \cup (P\cap Q)) \ominus P \subset X \ominus P$, $X$ 
is not a compatible program, a contradiction!
Thus $P\cap Q=\emptyset$ or $P\cap Q\subseteq X$.
In either case we have by set theory that
$P\setminus (P*Q)=P\setminus (X\cup Q)=P\setminus X$.
It can be shown in the same manner that 
$P\setminus (P*R)=P\setminus (X\cup R)=P\setminus X$.
Thus $P\setminus (P*Q)=P\setminus (P*R)$.
Again by the definition of slp-revision, we have 
$(P*Q)\setminus (P\cup Q)=(X\cup Q)\setminus (P\cup Q)=X\setminus P$.
Similarly 
$(P*R)\setminus (P\cup R)=(X\cup R)\setminus (P\cup R)=X\setminus P$.
Thus $(P*Q)\setminus (P\cup Q)=(P*R)\setminus (P\cup R)$.

For the other direction, suppose $*$ is a function that satisfies
(s$*$s), (s$*$c), (s$*$f), (s$*$rr), (s$*$er), (s$*$mr), and (s$*$u). 
We need to show $*$ is a slp-revision function.

Let $\gamma$ be defined as:
$$\gamma(P\updownarrow Q)=((P*Q)\cap P) \cup ((P*Q)\setminus Q)$$
for all $Q$.
It suffices to show $\gamma$ is a selection function for $P$ and
$P*Q=\gamma (P\updownarrow Q)\cup Q$.

Part 1:
For $\gamma$ to be a selection function, it must be a function.
Suppose $P\updownarrow Q= P\updownarrow R$.
Then (s$*$u) implies
$P\setminus (P*Q)=P\setminus (P*R)$
and 
$(P*Q)\setminus (P\cup Q)=(P*R)\setminus (P\cup R)$.
Since $P=(P\setminus (P*Q))\cup ((P*Q)\cap P)=
(P\setminus (P*R))\cup ((P*R)\cap P)$, 
$P\setminus (P*Q)=P\setminus (P*R)$ implies $(P*Q)\cap P=(P*R)\cap P$.
Thus $(P*Q)\setminus (P\cup Q)=(P*R)\setminus (P\cup R)$ implies
$((P*Q)\cap P)\cup ((P*Q)\setminus (P\cup Q))=((P*R)\cap P)\cup ((P*R)\setminus (P\cup R))$.
Then by set theory, we have 
$((P*Q)\cap P)\cup ((P*Q)\setminus Q)=((P*R)\cap P)\cup ((P*R)\setminus R)$.
Finally, it follows from the definition of $\gamma$ that
$\gamma(P\updownarrow Q)=\gamma(P\updownarrow R)$.

If $P\updownarrow Q=\emptyset$, then we have to show $\gamma(P\updownarrow Q)=P$.
$P\updownarrow Q=\emptyset$ implies $Q$ is m-inconsistent,
hence it follows from (s$*$f) that
$P*Q=P\cup Q$.
Then by the definition of $\gamma$,
$\gamma(P\updownarrow Q)=((P*Q)\cap P) \cup ((P*Q)\setminus Q)=
((P\cup Q)\cap P) \cup ((P\cup Q)\setminus Q)=P$.

If $P\updownarrow Q\not=\emptyset$, then we have to show
$\gamma(P\updownarrow Q)\in P\updownarrow Q$.
Since $P\updownarrow Q\not=\emptyset$, $Q$ is m-consistent.
Then (s$*$c) implies $P*Q$ is consistent.
Since
$\gamma(P\updownarrow Q)\cup Q=((P*Q)\cap P) \cup ((P*Q)\setminus Q)\cup Q=P*Q$,
$\gamma(P\updownarrow Q)\cup Q$ is consistent.
Assume there is $X$ s.t.\ $X\cup Q$ is consistent and
$X\ominus P\subset \gamma(P\updownarrow Q) \ominus P$.
Then we have three cases:

Case 1, there is $R$ s.t.\ 
$\emptyset\not=R\subseteq P \setminus \gamma(P\updownarrow Q)$, and $X=\gamma(P\updownarrow Q) \cup R$:
If $R\cap Q=\emptyset$, then 
since $\gamma(P\updownarrow Q)\cup Q=P*Q$, 
$R\cap (P*Q)=\emptyset$. Then it follows from
(s$*$rr) that $(P*Q) \cup R$ is inconsistent.
Since $X\cup Q =(P*Q) \cup R$, 
$X\cup Q$ is inconsistent, a contradiction!
If $R\cap Q\not=\emptyset$, then since $R\subseteq P$, $R\cap P\cap Q\not=\emptyset$.
Since (s$*$s) implies $Q\subseteq P*Q$, 
we have $Q\cap P \subseteq (P*Q)\cap P$, 
which implies $R\cap ((P*Q)\cap P)\not=\emptyset$.
Then since $((P*Q)\cap P)\subseteq \gamma(P\updownarrow Q)$,
$\gamma(P\updownarrow Q)\cap R\not=\emptyset$, a contradiction!
Thus $R\cap Q\not=\emptyset$ is an impossible case.

Case 2, there is $E$ s.t.\  $E\cap P=\emptyset$, $\emptyset \neq E\subseteq \gamma(P\updownarrow Q)$,
and $X= \gamma(P\updownarrow Q)\setminus E$:
Then $E\subseteq \gamma(P\updownarrow Q)\cup Q=P*Q$.
If $E\cap Q=\emptyset$, then (s$*$er) implies
$(P*Q)\setminus E$ is inconsistent.
Since $X\cup Q=\gamma(P\updownarrow Q)\setminus E\cup Q=
(P*Q)\setminus E$, $X\cup Q$ is inconsistent, a contradiction!
If $E\cap Q\not=\emptyset$, then $E\not\subseteq (P*Q)\setminus Q$.
Since $E\cap P=\emptyset$, we have $E\cap (P*Q)\cap P=\emptyset$.
Thus $E\not\subseteq ((P*Q)\cap P)\cup ((P*Q)\setminus Q)=\gamma(P\updownarrow Q)$,
a contradiction! Thus $E\cap Q\not=\emptyset$ is an impossible case.

Case 3, 
there are $R$ and $E$ s.t.\ 
$\emptyset\not=R\subseteq P$, $R\cap \gamma(P\updownarrow Q) =\emptyset$, 
$E\cap P=\emptyset$,
$\emptyset\neq E\subseteq \gamma(P\updownarrow Q)$,
and $X=(\gamma(P\updownarrow Q) \cup R) \setminus E$:
Then we can show as in Case 1 and 2 that $R\cap P*Q=\emptyset$ and $E\subseteq P*Q$.
If $R\cap Q=\emptyset$ and $E\cap Q=\emptyset$, 
then (s$*$mr) implies $((P*Q)\cup R)\setminus E$ is inconsistent.
Thus $X\cup Q= ((\gamma(P\updownarrow Q) \cup R) \setminus E)\cup Q=
((P*Q)\cup R)\setminus E$ is inconsistent, a contradiction!
Also we can show as in Case 1 and 2 that that
$R\cap Q=\emptyset$ and $E\cap Q=\emptyset$ are impossible cases.

Part 2: By set theory, 
$\gamma (P\updownarrow Q) \cup Q=
((P*Q)\cap P) \cup ((P*Q)\setminus Q)\cup Q=
((P*Q)\cap P) \cup (P*Q)=
P*Q$.

\qed

\bibliographystyle{named}
\bibliography{bibliography}

\end{document}